\documentclass[conference]{IEEEtran}
\IEEEoverridecommandlockouts
\usepackage{cite}
\usepackage{amsmath,amssymb,amsfonts}

\usepackage{graphicx}

\graphicspath{{figures/}}
\usepackage{multirow}  
\usepackage{array}     
\usepackage{subcaption}

\usepackage{textcomp}
\usepackage{xcolor}
\def\BibTeX{{\rm B\kern-.05em{\sc i\kern-.025em b}\kern-.08em
    T\kern-.1667em\lower.7ex\hbox{E}\kern-.125emX}}
\begin{document}
\raggedbottom

\title{Dynamic-TD3: A Framework for UAV Path Planning with adversarial environment}

\author{
Wentao Chen,
Jingtang Chen,
Mingjian Fu\textsuperscript{*},
Tiantian Li,
Youfeng Su,
Wenxi Liu,
and Yuanlong Yu
}

\maketitle

\begin{abstract}
Deep reinforcement learning (DRL) finds extensive application in autonomous drone navigation within complex, high-risk environments. However, its practical deployment faces a safety-exploration dilemma: soft penalty mechanisms encourage risky trial-and-error, while most constraint-based methods suffer degraded performance under sensor noise and intent uncertainty. We propose Dynamic-TD3, a physically enhanced framework that enforces strict safety constraints while maintaining maneuverability by modeling navigation as a Constrained Markov Decision Process (CMDP). This framework integrates an Adaptive Trajectory Relational Evolution Mechanism (ATREM) to capture long-range intentions and employs a Physically Aware Gated Kalman Filter (PAG-KF) to mitigate non-stationary observation noise. The resulting state representation drives a dual-criterion policy that balances mission efficiency against hard safety constraints via Lagrangian relaxation. In experiments with aggressive dynamic threats, this approach demonstrates superior collision avoidance performance, reduced energy consumption, and smoother flight trajectories.
\end{abstract}


\section{Introduction}

Unmanned aerial vehicles (UAVs) play a critical role in complex terrain reconnaissance, disaster relief, and national defense. Deep reinforcement learning (DRL) enables UAVs to autonomously perceive and make decisions in information-constrained, communication-disrupted environments, demonstrating exceptional real-time performance. However, the transition from simulation to actual deployment exposes the safety-exploration paradox. Existing DRL architectures struggle to handle uncertainties in extreme weather, dynamic obstacles, and unstructured environments, leading to unstable decision-making. Therefore, balancing mission efficiency with safety constraints under highly dynamic and noisy conditions remains a core challenge in autonomous navigation.

Existing continuous path planning methods rely on model-free reinforcement learning~\cite{1,3,4}, addressing safety through penalty-based approaches~\cite{Gao2024MaplessHRL,Huang2024GTRL}—which simplify safety requirements into sparse negative reward signals, forcing agents to identify safety boundaries through trial-and-error. However, this approach fails to address the nonlinear characteristics of high-frequency noise and dynamic obstacles, leading to policy volatility. Although Constrained Markov Decision Processes (CMDPs)~\cite{7,8,9} attempt to establish strict safety boundaries, they typically assume perfect state observability, neglecting the inherent uncertainty of noise and obstacle intent.

To address these challenges, this paper proposes Dynamic-TD3, a physically enhanced safety decision framework. We transform the UAV navigation problem into a CMDP and decouple safety costs from mission rewards by introducing a dual critic architecture with Lagrangian relaxation, ensuring rigid safety constraints while optimizing the policy. To overcome perception challenges, we design an adaptive trajectory correlation evolution mechanism to extract latent intentions and long-range spatio-temporal dependencies from unstructured threats. We further develop a physically augmented gated Kalman filter that leverages kinematic priors to suppress sensor anomalies and correct neural predictions. This enhanced physical-depth fusion mechanism enables agents to learn robust and efficient obstacle avoidance strategies in high-pressure adversarial environments.

The main contributions of this paper are summarized as follows:
\begin{itemize}
    \item We propose a constrained TD3 with a safety-cost critic and Lagrangian updates, separating safety costs from rewards to learn efficient, constraint-satisfying policies.

    \item We propose an adaptive trajectory evolution mechanism that captures intent manifolds and spatio-temporal dependencies, enhancing predictive perception to avoid anomalous or adversarial obstacles.

    \item We propose a physically augmented gated Kalman filter fusing kinematic priors with neural predictions for robust state estimation. 

    \item We build a 3D simulation platform with extreme dynamic threats for aggressive navigation evaluation, and experiments show our method consistently outperforms strong baselines under high-pressure conditions.

\end{itemize}

\section{Related Work}
\subsection{Online RL and Safety-Constrained Optimization}
In continuous control tasks such as drone navigation, policies must directly output continuous actions like velocity or acceleration. DDPG~\cite{lillicrap2016ddpg} implements off-policy learning via deterministic policy gradients, but suffers from training fluctuations under functional approximation errors; TD3~\cite{fujimoto2018td3} mitigates value overestimation through dual critic minimization, delayed updates, and target action smoothing; SAC~\cite{haarnoja2018sac} enhances exploration and robustness with a maximum entropy objective. Regarding sample efficiency, PER~\cite{schaul2016per} prioritizes sampling based on TD error, while I-TD3~\cite{ITD3} adapts this for TD3-style replays. However, neither addresses noisy observations or dynamic intent uncertainty, nor do they incorporate explicit safety constraints. Addressing CMDP's historical dependency, R2D2~\cite{kapturowski2019r2d2} enhances memory through recurrent structures but fails to predict dynamic obstacle intentions or perform physical denoising, hindering fundamental improvements in state reliability under noisy observations. Regarding safety, CMDP provides a framework for unified modeling of task rewards and safety costs. Lyapunov-safe reinforcement learning~\cite{chow2018lyapunov} offers structured constraint training, yet typically lacks perceptual enhancement support for noise observations and dynamic intent uncertainty, making stable deployment challenging in highly dynamic and noisy scenarios.

\subsection{RL for UAV Path Planning}
Reinforcement learning in drone path planning often centers on “reaching the target while avoiding obstacles,” directly outputting navigation commands or control quantities through continuous control. Hwangbo et al.~\cite{hwangbo2017quadrl} demonstrated a reinforcement learning-based quadcopter control method, validating DRL's feasibility for end-to-end policy learning under highly nonlinear dynamics; Kang et al.~\cite{kang2019gts} proposed a deep reinforcement learning framework integrating simulation and real-world data to achieve vision-driven autonomous flight while emphasizing cross-domain generalization; Kaufmann et al.~\cite{kaufmann2023nature} demonstrated a championship-level drone racing system, showcasing DRL's potential under high-speed, high-maneuverability conditions. Beyond dynamic obstacles, insufficient exploration due to sparse rewards remains a critical bottleneck in drone planning. Lv et al.~\cite{TD3ee} proposed Entropy Explorer (EE), constructing stable intrinsic rewards using state entropy and action entropy. Combined with TD3, the resulting TD3-EE framework improved planning success rates and optimized path quality.

\section{METHOD}

\begin{figure}[t]
\centering
\includegraphics[width=\linewidth]{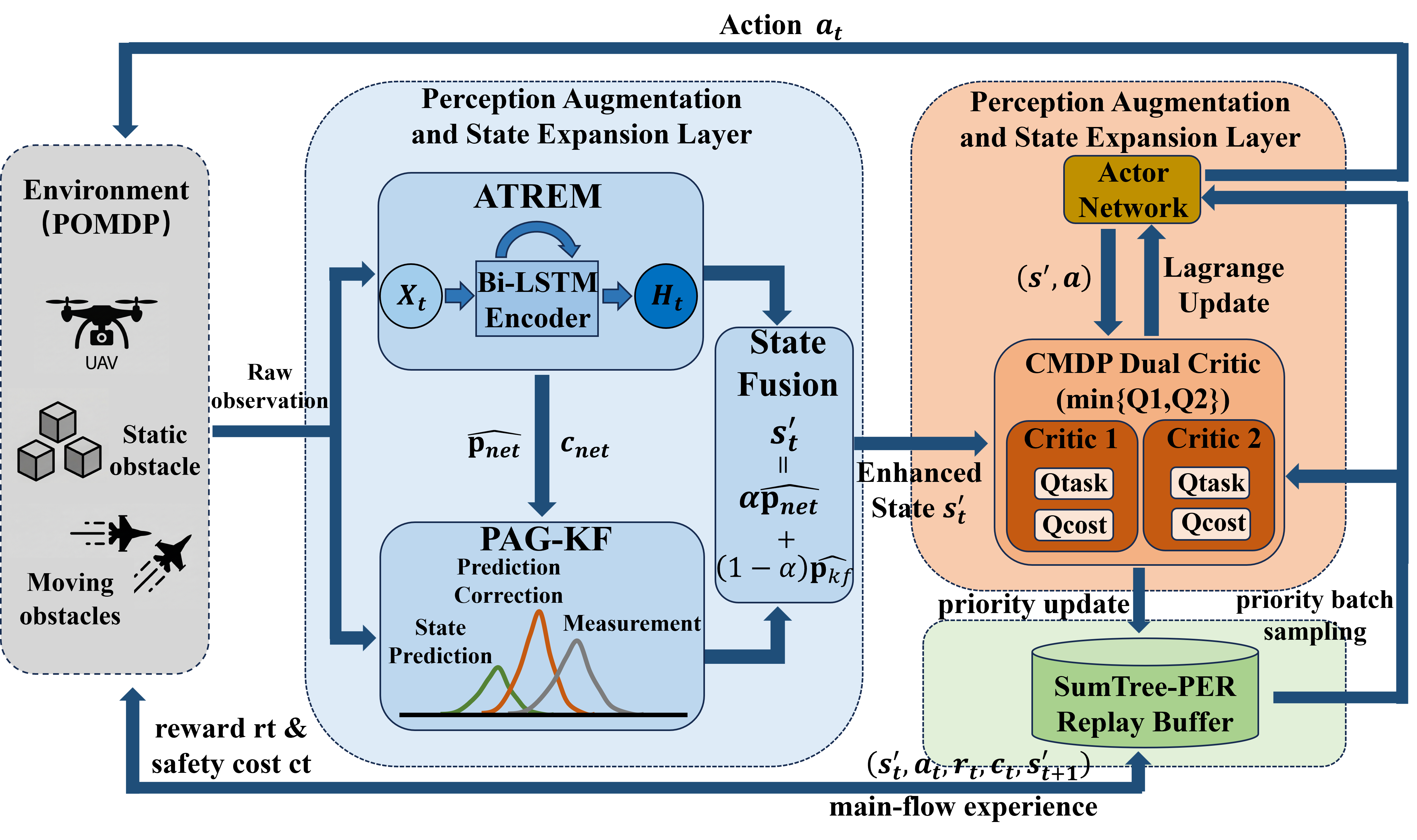}
\caption{Our proposed framework overview. Adaptive Trajectory Relational Evolution Mechanism(ATREM) predicts trajectory intent, while Physically Aware Gated Kalman Filter(PAG-KF) performs physically gated filtering for noise reduction and fuses enhanced states. The decision layer employs CMDP dual critic + Lagrangian constraint to safely output actions, with SumTree-PER enhancing training efficiency.}
\label{fig:Framework_Diagram}
\end{figure}

As shown in Figure \ref{fig:Framework_Diagram}, Dynamic-TD3 is introduced for CMDP drone planning to address challenges such as noisy observations, uncertain dynamic obstacle intentions, and the difficulty of ensuring safety with penalized rewards. ATREM predicts future trajectories and their confidence levels based on historical observations; PAG-KF performs gated fusion by treating these predictions as pseudo-observations alongside physical priors, yielding augmented states through state fusion. The decision layer employs a CMDP dual critic to simultaneously evaluate task benefits and safety costs, achieving adaptive constrained action output via Lagrange updates. The training module utilizes SumTree-PER~\cite{SumTree-PER} priority replay to enhance sample efficiency and stability.

\subsection{Simulation Environment}
This paper constructs a 3D simulation environment with a fixed height constraint. This design eliminates simple shortcuts in the vertical dimension, compelling agents to genuinely train their understanding of obstacles to resolve complex horizontal conflicts. Rather than achieving objectives by manipulating flight altitude, agents cannot resort to expedient means, preventing the model from learning the essence of obstacle avoidance. In experiments, agents must plan safe trajectories within kinematic constraints, simultaneously satisfying payload stability requirements while avoiding both dynamic and static obstacles.

\paragraph{CMDP under Partial Observability (Constrained POMDP)}
Since the maneuvering intent of dynamic obstacles is only partially observable, we formulate UAV navigation as a constrained POMDP.
Let $s_t\in\mathcal{S}$ be the latent state, $a_t\in\mathcal{A}$ the action, $\mathcal{T}(s_{t+1}\mid s_t,a_t)$ the transition model, $o_t\in\mathcal{O}$ the observation, and $\Omega(o_t\mid s_t)$ the observation model.
The agent receives task reward $r(s_t,a_t)$ and safety cost $c(s_t,a_t)$ with discount factor $\gamma\in(0,1)$, and aims to maximize return subject to $J_c(\pi)\le d$, where $J_c(\pi)=\mathbb{E}_\pi\!\left[\sum_{t=0}^{\infty}\gamma^t c(s_t,a_t)\right]$.
Under partial observability, the policy $\pi$ is conditioned on the observation history $h_t$ (or an equivalent belief state).

\paragraph{Dynamics and Action Space}
The environment evolves with discrete time steps $\Delta t$. The UAV state is $\mathbf{x}_t=[x_t,\,y_t,\,v_t,\,\psi_t]^\top$, where $(x_t,y_t)$ represents position, $v_t$ denotes velocity, and $\psi_t$ indicates heading angle. Actions are continuous control variables $\mathbf{a}_t=[u_t,\,\omega_t]^\top$, representing longitudinal acceleration and angular velocity, respectively. The policy outputs normalized actions $\tilde{\mathbf{a}}_t\in[-1,1]^2$ and linearly maps them to the physical range. The state transition model for the UAV is: 
\begin{equation}
\begin{aligned}
x_{t+1} &= x_t + v_t\cos(\psi_t)\Delta t,\\
y_{t+1} &= y_t + v_t\sin(\psi_t)\Delta t,\\
v_{t+1} &= \mathrm{clip}(v_t+u_t\Delta t, v_{\min}, v_{\max}),\\
\psi_{t+1} &= \mathrm{clip}(\psi_t+\omega_t\Delta t, \psi_{\min}, \psi_{\max}),
\end{aligned}
\label{eq:dynamics}
\end{equation}
where $u_t$ and $\omega_t$ are control inputs, $v_{\min}, v_{\max}$ and $\psi_{\min}, \psi_{\max}$ are velocity and heading angle constraints, and $\mathrm{clip}(\cdot)$ is the clipping operator ensuring variables satisfy constraints.

\paragraph{Observation Space (29 Dimensions)}
This observation consists of three parts: (1) UAV self-state: 4 dimensions comprising 2D position and 2D velocity components; (2) Target information: target position, distance to target, and unit direction vector pointing toward the target, totaling 5 dimensions; (3) Local environment information: positions of $3$ static obstacles totaling 6 dimensions, positions of $5$ dynamic obstacles totaling 10 dimensions, and the observed velocity and acceleration of the nearest dynamic obstacle, totaling 4 dimensions.

\paragraph{Termination Conditions and Collision Constraints}
Rounds terminate upon reaching the target neighborhood, encountering a collision, or exceeding the maximum number of steps. A circular safety region approximates geometric constraints, with termination conditions defined as:
\begin{equation}
\|P_t-P_G\|_2 \le R_G,\quad
\|P_t-P^O_i\|_2 \ge R^O_i,\quad
\|P_t-P^E_{j,t}\|_2 \ge R_j^E.
\label{eq:terminal_collision}
\end{equation}
where $P_t$ denotes the drone's position, $P_G$ denotes the target's position, $R_G$ denotes the reach radius, $R^O_i$ and $R_j^E$ denote the safety radii for static and dynamic obstacles.

\paragraph{Reward Function}
To simultaneously incentivize reaching the target, avoiding collisions, and improving trajectory efficiency, this paper employs an additive reward design:
\begin{equation}
r_t = r_t^{\mathrm{goal}} + r_t^{\mathrm{crash}} + r_t^{\mathrm{shape}}
+ r_t^{\mathrm{edge}} + r_t^{\mathrm{risk}} - c_{\mathrm{step}} .
\label{eq:reward_total}
\end{equation}
The total reward comprises goal-reaching $r_t^{\mathrm{goal}}$, collision $r_t^{\mathrm{crash}}$, shape-shaping $r_t^{\mathrm{shape}}$, boundary-crossing $r_t^{\mathrm{edge}}$, risk penalty $r_t^{\mathrm{risk}}$, and step-size penalty $c_{\mathrm{step}}$. The goal-reached and collision components provide rewards or penalties via indicator functions upon goal attainment or collision. Distance shaping guides the agent toward the goal, while edge crossing and risk penalties respectively penalize boundary violations and proximity to obstacles. The step size penalty suppresses aimless wandering and enhances arrival efficiency.

\subsection{Perception Enhancement and Adaptive Noise Suppression}
Sensor observations in dynamic UAV navigation often contain non-stationary noise and sudden anomalies. To address this issue, we propose the PAG-KF as a perception front-end rectifier, combining physical prior knowledge to perform online correction and anomaly suppression on the observation stream.

\textbf{Prediction (Physical Benchmark).}
We adopt a constant-acceleration model to propagate the state and uncertainty.
The state is $\mathbf{x}_t=[p_x, p_y, v_x, v_y, a_x, a_y]^\top$, and the covariance is $P_t$.
The prediction step is~\cite{KF}:
\begin{equation}
\mathbf{x}_{t|t-1} = F\mathbf{x}_{t-1|t-1}, \qquad
P_{t|t-1} = F P_{t-1|t-1} F^\top + Q,
\end{equation}
where $F$ is the state transition matrix and $Q$ is the process noise covariance.

\textbf{Innovation Statistics (Adaptive Noise).} For observation $\mathbf{z}_t$, compute the innovation residual $\tilde{\mathbf{y}}_t = \mathbf{z}_t - H \mathbf{x}_{t|t-1}$, with covariance $S_t = H P_{t|t-1} H^\top + R_t$. Based on historical innovation window statistics (e.g., mean and standard deviation), PAG-KF adaptively adjusts the measurement noise covariance $R_t$.

\textbf{Gated Arbitration (Outlier Suppression).} Observation bias is measured via Mahalanobis distance $d_t^2 = \tilde{\mathbf{y}}_t^\top S_t^{-1} \tilde{\mathbf{y}}_t$. If $d_t^2 > \eta$, the observation is deemed anomalous, and the Kalman gain is adjusted using the gating coefficient $\gamma_t$:
\begin{equation}
\gamma_t =
\begin{cases}
\frac{\eta}{d_t^2}, & d_t^2 > \eta,\\
1, & d_t^2 \le \eta.
\end{cases}
\end{equation}

In summary, PAG-KF rapidly fuses information during reliable observations while suppressing gains during anomalies, enhancing perception robustness and decision stability.

\subsection{Adaptive Trajectory-Related Evolution Mechanism}
To address nonlinear and noisy disturbances in intent-driven maneuvers, we propose ATREM to achieve stable, forward-looking intent representation.
Given an input sequence $X=[x_1,\dots,x_T]\in\mathbb{R}^{T\times 6}$, we first perform manifold projection and noise suppression:
\begin{equation}
z_t = W_{e2}\,\psi\!\big(\mathrm{LN}(W_{e1}x_t + b_{e1})\big) + b_{e2},\ \ t=1,\dots,T,
\end{equation}
where $\mathrm{LN}(\cdot)$ denotes layer normalization and $\psi(\cdot)$ is a nonlinear activation function (e.g., GELU),
$Z=[z_1,\dots,z_T]\in\mathbb{R}^{T\times d}$.

Contextual features are then extracted using bidirectional temporal encoding~\cite{Bi-LSTM}, and key moments/interactions are aggregated via multi-head self-attention:
\begin{equation}
H=\mathrm{BiLSTM}(Z),\quad A=\mathrm{MHAttn}(H).
\end{equation}
where $H$ encodes spatio-temporal context and $A$ is the attention-aggregated relation representation.

The evolutionary gate mechanism adapts between global trends and transient states:
\begin{equation}
F = \sigma(g)\odot \mathrm{Mean}_t(A) + (1-\sigma(g))\odot A_{\text{last}},
\end{equation}
where $\sigma(g)$ is a learnable gating function and $\mathrm{Mean}_t(\cdot)$ denotes temporal pooling. This enables the model to prioritize long-term trends while rapidly responding to instantaneous actions, ensuring robustness in intention representation for safety-critical decisions.

\subsection{Safety-Constrained Control and Efficient Learning Mechanism}

\textbf{ATREM and PAG-KF Cooperative Perception Fusion.}
The algorithm first performs noise filtering and enhancement. To address the uncertainty introduced by dynamic obstacles, we establish a collaboration between ATREM and PAG-KF: ATREM predicts future positions from the raw observation tensor $\mathcal{O}_t$ and provides confidence levels:
\begin{equation}
[\hat{\mathbf{p}}_{\text{net}},\,c_{\text{net}}]
= \Phi_{\text{ATREM}}(\mathcal{O}_t),
\end{equation}
where $\hat{\mathbf{p}}_{\text{net}}$ is the predicted position vector and $c_{\text{net}}$ is its confidence level. Subsequently, PAG-KF treats $\hat{\mathbf{p}}_{\text{net}}$ as a pseudo-measurement and performs a physically consistent correction. The measurement noise covariance is set according to the baseline noise $\sigma_b$ and scaling factor $k$:
\begin{equation}
R_t = (\sigma_b k)^2 I,\qquad
\hat{p}_{\text{kf}}=\mathrm{PAG\text{-}KF}(\hat{p}_{\text{net}}).
\end{equation}
Finally, “predictive information” and the “physically smoothed estimate” are fused into an augmented state:
\begin{equation}
s'_t=\alpha\,\hat{\mathbf{p}}_{\text{net}} + (1-\alpha)\,\hat{\mathbf{p}}_{\text{kf}},
\end{equation}
where $\alpha\in[0,1]$ is the fusion weight (constant or adaptive with $c_{\text{net}}$), and $s'_t$ is input to the downstream safety decision layer.

\textbf{CMDP Dual Critic.}
To satisfy the CMDP constraints, we employ a TD3-style dual critic $i\in\{1,2\}$, where each critic comprises a task value head and a cost value head:
$Q^{(i)}_{\text{task}}(s,a)$ evaluates rewards, while $Q^{(i)}_{\text{cost}}(s,a)$ assesses safety costs (e.g., proximity to threat prediction trajectories).
The objective values are respectively:
\begin{equation}
y_{\text{task}} = r_t + \gamma (1-d_t)\min_{i=1,2} Q^{\prime(i)}_{\text{task}}(s'_{t+1}, a'_{\text{target}}),
\end{equation}
\begin{equation}
y_{\text{cost}} = c_t + \gamma(1-d_t)\max_{i=1,2} Q_{\text{cost}}^{\prime(i)}(s'_{t+1},a'_{\text{target}}).
\end{equation}
where $a'_{\text{target}}$ denotes the smoothed action of the target Actor, and $d$ is the termination flag; the total loss is $L_C=L_{\text{task}}+L_{\text{cost}}$ (both heads use MSE).

To maximize task reward while satisfying safety constraints, the Actor network achieves constraint optimization by maximizing task value and penalizing cost~\cite{Safe-RL}:
\begin{equation}
\begin{aligned}
\nabla_{\phi} J(\phi)
\approx\;
\mathbb{E}_{s \sim \mathcal{D}}
\Big[
\nabla_{\phi}\mu_{\phi}(s)
\Big(
\nabla_{a} Q_{\text{task},\theta_1}(s,a)
\\
\qquad\qquad
-\lambda \nabla_{a} Q_{\text{cost},\psi}(s,a)
\Big)
\Big]_{a=\mu_{\phi}(s)} .
\end{aligned}
\label{eq:actor_update}
\end{equation}
where $\lambda$ is the Lagrange multiplier, used to adaptively balance reward and safety costs.

\textbf{SumTree-PER Efficient Sampling.}
To avoid the linear scanning overhead of traditional priority sampling, we maintain priorities using SumTree, achieving $O(\log N)$ sampling and updates. The priority of each experience is:
\begin{equation}
P(i)=\frac{p_i^\alpha}{\sum_j p_j^\alpha}.
\end{equation}
where $\alpha$ controls the priority strength. This mechanism focuses training on high-value samples, accelerating convergence and meeting efficient learning requirements.

\section{EXPERIMENTS}
\label{sec:experiments}

\subsection{Experimental Setup}
\label{subsec:exp_setup}
We evaluated the performance of Dynamic-TD3 against mainstream reinforcement learning algorithms (DDPG, SAC, TD3) and their variants (I-TD3, TD3\_EE) within a unified dynamic obstacle avoidance simulation environment. Scenarios ranged from low-density obstacles (S1+D1) to high-density dynamic obstacles (S3+D5). The state space comprised the drone's position, velocity, and relative positions of enemy aircraft and obstacles, while the action space consisted of continuous control variables (acceleration and angular acceleration). Each episode had a maximum of 500 steps, using the Adam optimizer with a learning rate of 1e-4 and a batch size of 64. Experiments employed a fixed random seed to ensure reproducibility. All experiments were conducted on an NVIDIA GeForce RTX 4070.

\textbf{Simulation Scenario.} The scenario includes one autonomous drone, $l$ static obstacles, and $m$ dynamic obstacles. Each round, the drone's initial state and target position are randomly sampled, while the initial obstacle distribution is randomly generated. The agent must successfully reach the target neighborhood within a finite number of steps, avoiding collisions and adhering to boundary constraints.

\subsection{Comparison Results}

\subsubsection{Training Analysis}

\begin{figure}[htbp]
\centering
\begin{subfigure}[b]{0.24\textwidth}
\centering
\includegraphics[width=\linewidth]{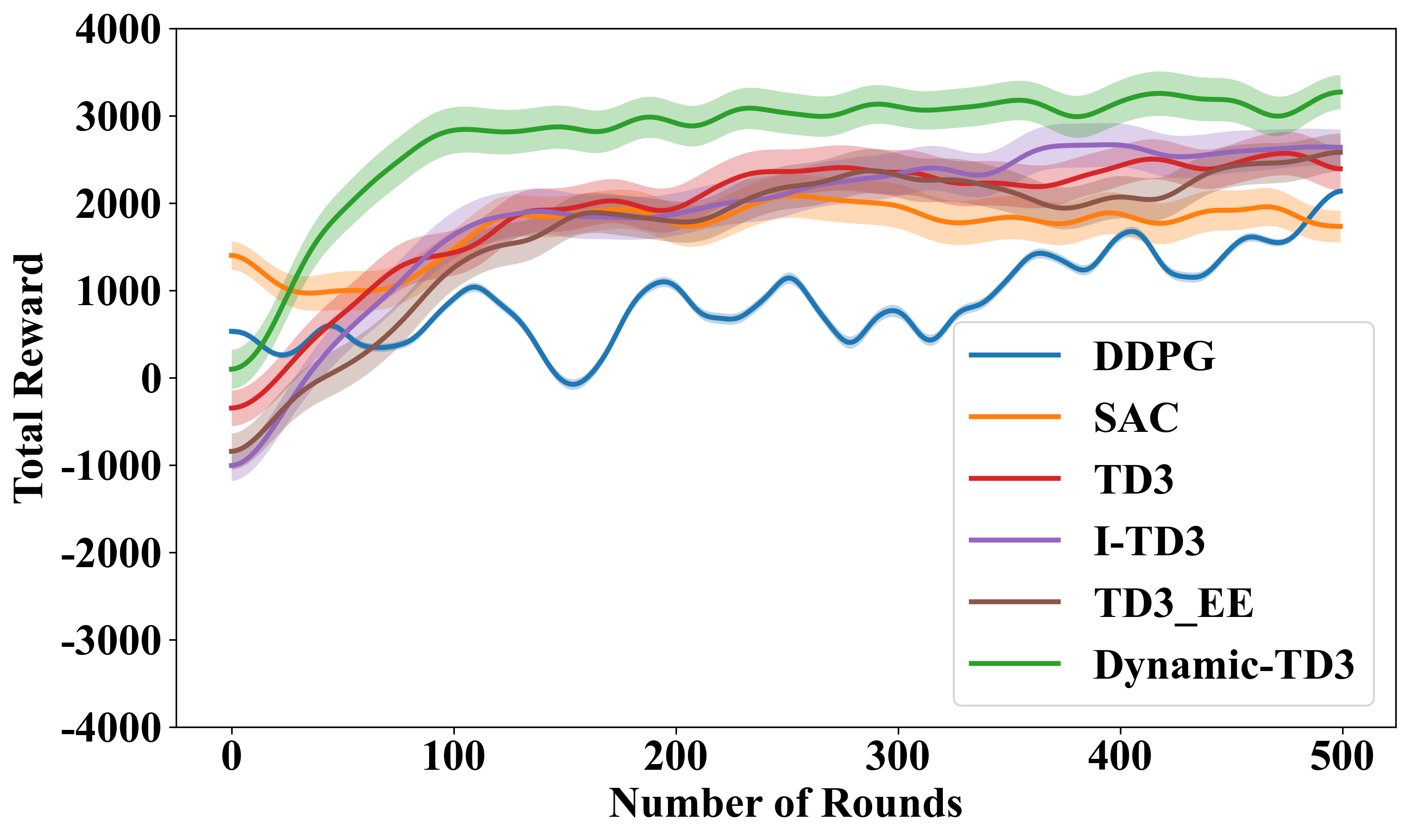}
\caption{Total Reward Comparison}
\label{total_reward_compare}
\end{subfigure}
\hfill 
\begin{subfigure}[b]{0.24\textwidth}
\centering
\includegraphics[width=\linewidth]{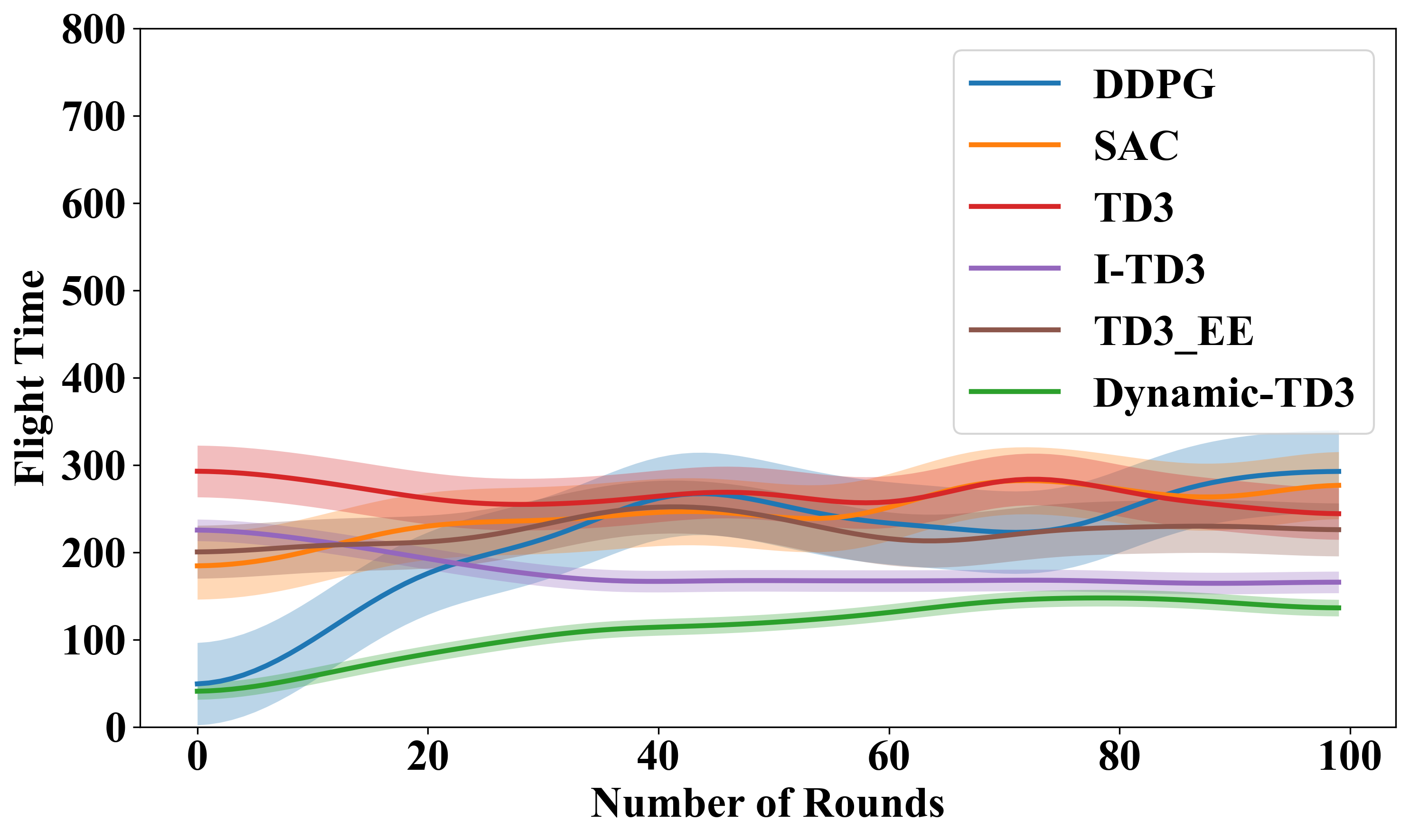}
\caption{Flight Time Comparison}
\label{flight_time_compare}
\end{subfigure}
\caption{This figure demonstrates the comparison of total training reward and flight time across different algorithms. Dynamic-TD3 performs exceptionally well on both metrics, significantly outperforming other algorithms.}
\label{fig:comparison_experiment}
\end{figure}

Figure \ref{total_reward_compare} illustrates the total reward evolution curves of Dynamic-TD3 and baseline algorithms, reflecting their sample efficiency and robustness in non-stationary noisy environments.

Throughout training, DDPG exhibits significant fluctuations and fails to converge stably, primarily due to its deterministic policy's inability to effectively handle noisy observations. SAC explores rapidly under the maximum entropy mechanism, but its reward stagnates after approximately 200 episodes, indicating an overly conservative balance between safety and exploration that prevents escaping local optima.

In contrast, Dynamic-TD3 demonstrates significant performance advantages:
First, it exhibits high sample efficiency, rapidly outperforming other algorithms early in training, validating the effectiveness of the SumTree-PER mechanism. Second, it achieves optimal asymptotic performance, with rewards exceeding 3000 points at training completion, benefiting from the dynamic obstacle intent manifold extracted by ATREM. Furthermore, the introduction of PAG-KF effectively suppresses observation anomalies, ensuring stable policy updates.

Figure \ref{flight_time_compare} reveals an abnormally high flight time for DDPG, indicating excessive conservatism. While traditional TD3 and its variants exhibit stable performance, they incur longer flight times (200–250 seconds). Dynamic-TD3 consistently achieves the lowest flight time (approximately 150 seconds), demonstrating the superiority of decoupling physical perception from decision-making. PAG-KF eliminates noise fluctuations through smooth state estimation, while ATREM's intent prediction enables the agent to proactively avoid obstacles, achieving more efficient and safe navigation.

\begin{table*}[t]
\caption{Performance Comparison of Different Algorithms}
\label{tab:comparison_boxed}
\centering
\resizebox{0.8\textwidth}{!}{
\begin{tabular}{|c|c|c|c|c|c|c|c|}
\hline
\multirow{2}{*}{\textbf{Scenario}} & \multirow{2}{*}{\textbf{Metric}} & \multicolumn{6}{c|}{\textbf{Algorithms}} \\
\cline{3-8}
& & \textbf{\textit{DDPG}} & \textbf{\textit{SAC}} & \textbf{\textit{TD3}} & \textbf{\textit{I-TD3}} & \textbf{\textit{TD3\_EE}} & \textbf{\textit{Dynamic\_TD3}} \\
\hline
\multirow{4}{*}{\shortstack{Static 1\\Dynamic 1}} 
& Success Rate $\uparrow$ & 80\% & 88\% & 90\% & 91\% & 86\% & \textbf{92\%} \\ \cline{2-8}
& Flight Time (s) $\downarrow$ & 353.92 & 182.88 & 270.05 & 246.98 & 255.00 & \textbf{171.46} \\ \cline{2-8}
& Trajectory (m) $\downarrow$ & 328.82 & \textbf{151.64} & 171.13 & 170.30 & 188.96 & 269.21 \\ \cline{2-8}
& Energy Cons. $\downarrow$ & 352.74 & 132.63 & 168.85 & 216.53 & 291.14 & \textbf{129.00} \\ \hline

\multirow{4}{*}{\shortstack{Static 3\\Dynamic 2}} 
& Success Rate $\uparrow$ & 70\% & 72\% & 77\% & 79\% & 76\% & \textbf{81\%} \\ \cline{2-8}
& Flight Time (s) $\downarrow$ & 209.60 & 174.72 & 182.75 & 265.72 & 320.67 & \textbf{159.53} \\ \cline{2-8}
& Trajectory (m) $\downarrow$ & 197.94 & \textbf{156.40} & 158.20 & 204.21 & 195.27 & 259.91 \\ \cline{2-8}
& Energy Cons. $\downarrow$ & 230.96 & 121.83 & 158.15 & 323.79 & 279.95 & \textbf{114.31} \\ \hline

\multirow{4}{*}{\shortstack{Static 3\\Dynamic 3}} 
& Success Rate $\uparrow$ & 65\% & 67\% & 73\% & 75\% & 71\% & \textbf{76\%} \\ \cline{2-8}
& Flight Time (s) $\downarrow$ & 239.00 & 180.41 & 190.59 & 274.29 & 259.53 & \textbf{154.52} \\ \cline{2-8}
& Trajectory (m) $\downarrow$ & 183.14 & \textbf{149.40} & 177.13 & 213.76 & 166.75 & 258.10 \\ \cline{2-8}
& Energy Cons. $\downarrow$ & 191.40 & 142.76 & 160.67 & 305.63 & 256.35 & \textbf{135.56} \\ \hline

\multirow{4}{*}{\shortstack{Static 3\\Dynamic 4}} 
& Success Rate $\uparrow$ & 57\% & 62\% & 68\% & 67\% & 69\% & \textbf{71\%} \\ \cline{2-8}
& Flight Time (s) $\downarrow$ & 216.50 & 201.22 & 194.59 & 180.04 & 232.12 & \textbf{136.77} \\ \cline{2-8}
& Trajectory (m) $\downarrow$ & 205.19 & \textbf{133.81} & 175.19 & 144.68 & 155.89 & 238.32 \\ \cline{2-8}
& Energy Cons. $\downarrow$ & 292.98 & 154.58 & 167.83 & 282.52 & 173.82 & \textbf{151.78} \\ \hline

\multirow{4}{*}{\shortstack{Static 3\\Dynamic 5}} 
& Success Rate $\uparrow$ & 45\% & 59\% & 60\% & 64\% & 65\% & \textbf{67\%} \\ \cline{2-8}
& Flight Time (s) $\downarrow$ & 276.70 & 238.03 & 247.42 & 161.77 & 228.69 & \textbf{139.79} \\ \cline{2-8}
& Trajectory (m) $\downarrow$ & 251.84 & 151.08 & 156.06 & \textbf{130.48} & 175.40 & 252.08 \\ \cline{2-8}
& Energy Cons. $\downarrow$ & 261.35 & 199.24 & 215.23 & 196.23 & 192.75 & \textbf{168.41} \\ \hline

\multicolumn{8}{l}{$^{\mathrm{a}}$Bold indicates the best performance.}
\end{tabular}
}
\end{table*}

\subsubsection{Testing Analysis}
\label{subsec:comparative}
We employ task success rate, flight time, energy consumption, and trajectory length as evaluation metrics. Table~\ref{tab:comparison_boxed} illustrates the performance of each algorithm across varying difficulty scenarios.

As environmental complexity increases, DDPG and SAC perform poorly in highly dynamic settings. While SAC achieves shorter paths and lower energy consumption, its safety is compromised, with a success rate of only {62\%} due to overly aggressive strategies driven by its reward mechanism. TD3 performs well in low-to-medium difficulty scenarios, but in the {S3+D5} scenario, its success rate drops to {60\%} and flight time increases to {247.42s}, indicating it adopts conservative strategies in complex environments, leading to reduced efficiency.

Dynamic-TD3 excels in the {S3+D5} scenario, achieving a success rate of {67\%} and reducing flight time to {139.79s}—an improvement of approximately {43.5\%} over traditional TD3. By adopting a “space-for-time” strategy, Dynamic-TD3 avoids stagnation and maintains smooth path planning, thereby lowering both flight time and energy consumption.

\begin{table*}[t]
\caption{Ablation Study of Different Components}
\label{tab:ablation_study}
\centering
\resizebox{0.8\textwidth}{!}{
\begin{tabular}{|c|c|c|c|c|c|c|c|}
\hline
\textbf{Index} & \textbf{Safe-decision} & \textbf{ATREM} & \textbf{PAG-KF} & \textbf{Flight Time (s) $\downarrow$} & \textbf{Energy Cons. $\downarrow$} & \textbf{Trajectory (m) $\downarrow$} & \textbf{SR $\uparrow$} \\ \hline

A & & & & 247.42 & 215.23 & \textbf{156.06} & 60\% \\ \hline
B & & $\checkmark$ & $\checkmark$ & 179.71 & 176.17 & 312.35 & 65\% \\ \hline
C & $\checkmark$ & & & 235.42 & 219.02 & 257.06 & 62\% \\ \hline
D & $\checkmark$ & $\checkmark$ & & 159.73 & 180.71 & 246.22 & 64\% \\ \hline
E & $\checkmark$ & & $\checkmark$ & 153.72 & 252.75 & 239.34 & 63\% \\ \hline
F & $\checkmark$ & $\checkmark$ & $\checkmark$ & \textbf{139.79} & \textbf{168.41} & 252.08 & \textbf{67\%} \\ \hline

\multicolumn{8}{l}{$^{\mathrm{a}}$Bold indicates the best performance; $\checkmark$ indicates the component is included.}
\end{tabular}
}
\end{table*}

\subsection{Ablation Study}

\subsubsection{Training Analysis}

\begin{figure}[htbp]
\centering
\begin{subfigure}[b]{0.24\textwidth}
\centering
\includegraphics[width=\linewidth]{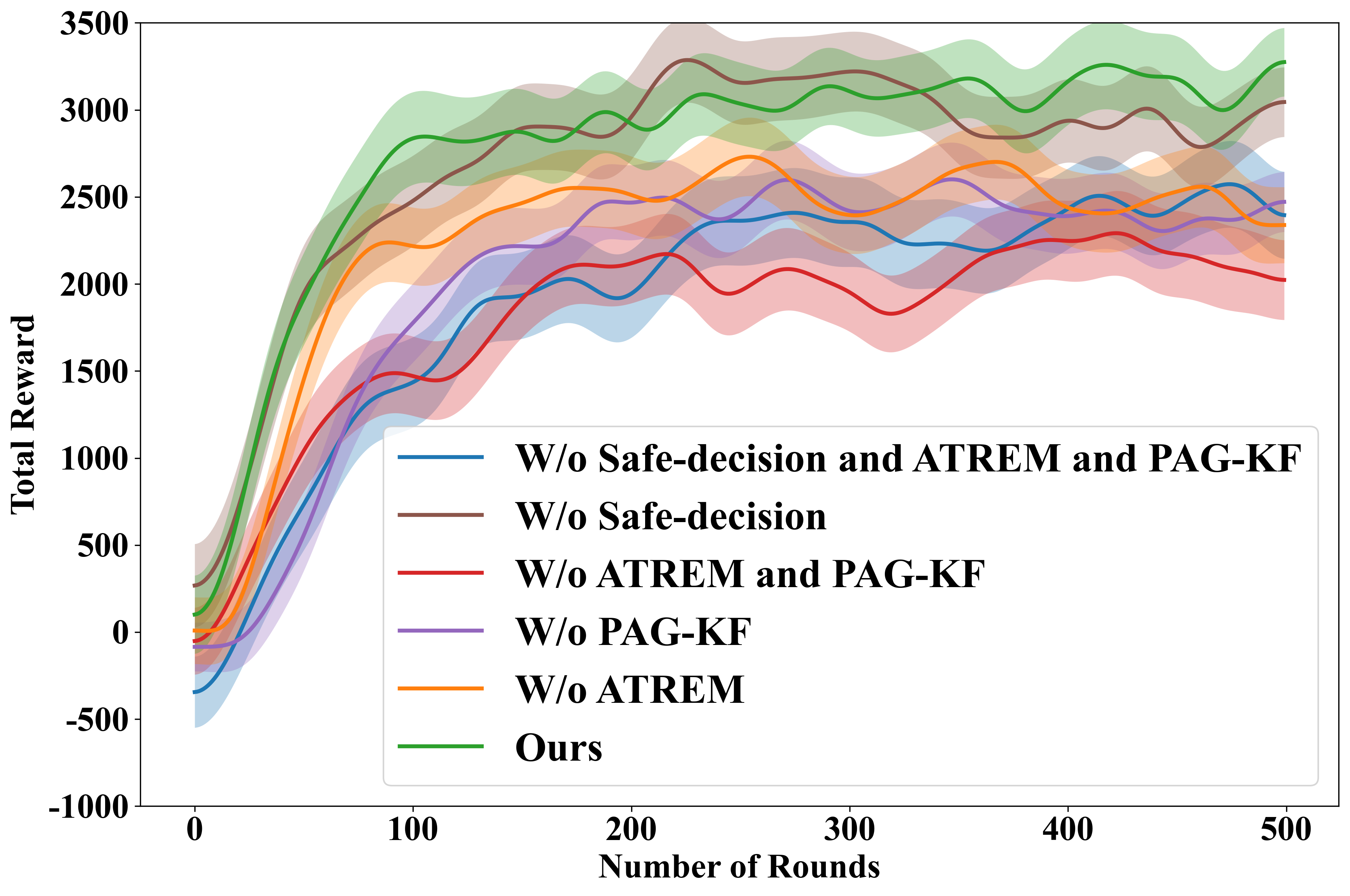}
\caption{Total Training Reward Curve}
\label{total_reward_ablation}
\end{subfigure}
\hfill 
\begin{subfigure}[b]{0.24\textwidth}
\centering
\includegraphics[width=\linewidth]{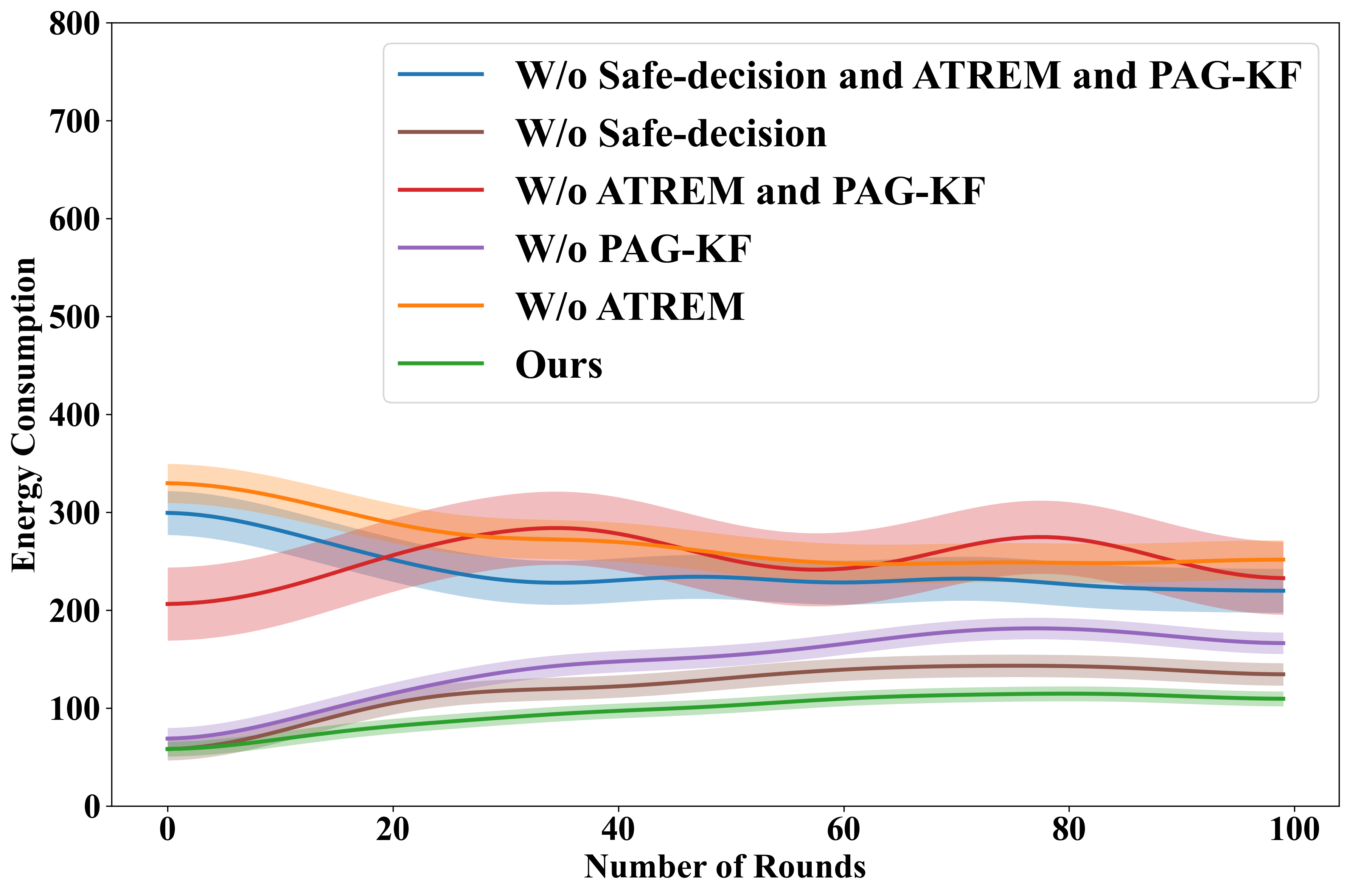}
\caption{Energy Consumption}
\label{energy_ablation}
\end{subfigure}

\caption{Ablation experiment results. The impact of each component on total training reward and test flight duration demonstrates the complete method's advantages in both reward and efficiency.}
\label{fig:ablation_experiment}
\end{figure}

To validate the effectiveness of the core components (ATREM, PAG-KF, Safe-decision) in Dynamic-TD3, we analyzed the training dynamics and flight performance of different variant models.

As shown in Figure \ref{total_reward_ablation}, the full model (Ours) demonstrates optimal sample efficiency and progressive performance. Benefiting from the SumTree-PER efficient sampling strategy, it rapidly increases during early training and converges to a high-reward range after approximately 200 iterations. In contrast, removing ATREM (W/o ATREM) reduces policy learning efficiency, while removing PAG-KF (W/o PAG-KF) leads to unstable state estimation, causing the final reward to plateau between 2400 and 2600.

Figure \ref{energy_ablation} demonstrates the complete model's excellent energy efficiency, ultimately converging to a low energy level (approximately 100). W/o PAG-KF results in higher energy consumption and frequent fluctuations, indicating its effective noise filtering. While removing the safety decision module (W/o Safe-decisoin) yields better rewards, its energy consumption peaks at 200, revealing the agent's tendency toward aggressive strategies.

In summary, Dynamic-TD3 achieves a balance between task efficiency and physical robustness by integrating CMDP constraints with physical perception, thereby avoiding the potential risks of solely pursuing reward maximization.

\subsubsection{Testing Analysis}
To validate the effectiveness of each key module in {Dynamic-TD3}, we conducted ablation experiments. Table~\ref{tab:ablation_study} shows performance changes after removing Safe-decision, ATREM, and PAG-KF. Results indicate that removing any module leads to performance degradation, particularly in task success rate, flight time, and energy consumption.

Removing ATREM increases flight time from {159.73s} to {235.42s} (a {47.4\%} increase), indicating ATREM's critical role in enhancing decision efficiency and reducing stagnation. Integrating PAG-KF improved success rate to {67\%} while reducing flight time to {139.79s}, validating PAG-KF's contribution to state estimation accuracy and system stability. Retaining only PAG-KF decreased flight time but increased energy consumption to {252.75}, highlighting the importance of their synergistic interaction.

Removing Safe-decision reduced the success rate to {65\%} while other metrics also declined, proving the critical role of safety constraints in ensuring flight stability and avoiding high-risk maneuvers.

\section{CONCLUSION}
We propose Dynamic-TD3, a robust and safe navigation framework for UAVs operating in highly noisy and dynamically threatening environments. By modeling the navigation problem as a CMDP, we successfully decouple safety constraints from mission efficiency. Two perception innovations are introduced: PAG-KF fuses motion priors to adaptively suppress non-stationary noise, yielding stable state estimates; ATREM extracts long-temporal-spatial interaction features from historical trajectories of dynamic obstacles, enabling early characterization and prediction of maneuvering intent. Extensive high-fidelity simulations demonstrate that Dynamic-TD3 significantly outperforms existing benchmark algorithms in success rate, energy efficiency, and trajectory smoothness. These results validate the effectiveness of integrating physical perception with constraint optimization for achieving safe and efficient autonomy in unstructured environments.

\bibliographystyle{IEEEbib}
\bibliography{icme2026references}

\vspace{12pt}

\end{document}